%% file: main.tex
\documentclass{article} 
\usepackage{iclr2027_conference,times}

\usepackage{hyperref}
\usepackage{url}

\usepackage{amsmath,amssymb,amsthm}
\usepackage{booktabs}
\usepackage{graphicx}
\usepackage{tikz}
\usetikzlibrary{arrows.meta,positioning,fit,decorations.pathreplacing,calc}
\usepackage{xspace}
\makeatletter\newcommand\captionof[1]{\def\@captype{#1}\caption}\makeatother

\newcommand{\spd}{\textsc{SPD}\xspace}
\DeclareRobustCommand{\rev}[1]{{\color{black}#1}}
\DeclareMathOperator*{\argmax}{arg\,max}
\newcommand{\Reals}{\mathbb{R}}
\newcommand{\Perm}{\Pi}
\newcommand{\Mmat}{\mathbf{M}}
\newcommand{\Pteach}{\mathbf{P}^{\mathrm{teacher}}}
\newcommand{\Smat}{\mathbf{S}}
\newcommand{\hvec}{\mathbf{h}}
\newcommand{\speedup}{$64\times$}

\title{SPD: Single Pass Decoding for Generative Reranking}

\author{%
Emil Laftchiev, Prachi Agrawal, Moe Kayali, \&  Bixing Yan \\
Meta Platforms, Inc. \\
\texttt{\{emillaftchiev,praa,kayali,bixing\}@meta.com} \\
\And
 Qi Xu, Zijie Lei, Chen Qiu, Zhi Hua, Ke Li \& Luke Simon\\
Meta Platforms, Inc. \\
}

\iclrfinalcopy 

\begin{document}
\maketitle

\begin{abstract}
Large language models (LLMs) achieve state-of-the-art generative ranking quality, but
the ranking they produce must be \emph{decoded}, and autoregressive decoding spends one
sequential forward pass per emitted token. We observe that the only tokens a ranker must
emit are the $N$ ordinal values, corresponding to the number of positions available to rank, naming the items in ranked order, and that this narrow,
permutation-structured output format admits decoding strategies which are much more efficient than left-to-right generation. We introduce \spd (Single Pass Decoding), a format-specialized
decoding strategy that decodes all $N$ ordinals in $O(1)$ forward passes. \spd reads an
$N \times K$ item-position score matrix off the LLM's prefill hidden states with a
lightweight self-attention head, then decodes the ordinals as the optimal bipartite
assignment of that matrix via the Hungarian algorithm, yielding a valid permutation by
construction rather than by repair, without enumerating candidate prefixes or re-querying the LLM. Through a systematic study of training signals and
backbone adaptation, we show that LoRA-based fine-tuning combined with autoregressive LLM
ranking distillation reaches \textbf{28\,ms} end-to-end inference, \rev{$3.1\times$ faster than a same-size autoregressive teacher without reasoning ($64\times$ faster than the teacher with reasoning)} while maintaining ranking quality on par. \rev{On four public benchmarks, a $0.6$B \spd student stays within 4\% of its 32B teacher's AUC at $44$--$88\times$ lower latency. An autoregressive student trained on the same teacher reaches similar quality but is $14$--$41\times$ slower, so distillation supplies the quality and single-pass decoding supplies the speed.} We provide a complete ablation decomposing the contributions of architecture, training signal, and backbone adaptation\rev{, and compare against five efficient rerankers}. Our framework connects generative ranking to combinatorial optimization, opening a path toward other $O(1)$-decode mechanisms for real-time ranking.
\end{abstract}

\section{Introduction}
\label{sec:intro}

Generative ranking, producing a complete ordering of $N$ candidates jointly rather
than scoring each independently, is critical for recommendation, advertising, and
search. Autoregressive LLMs have recently emerged as powerful generative rankers,
achieving superior quality by jointly reasoning over all candidates in context
\citep{DBLP:conf/emnlp/0001YMWRCYR23}. The output of such a ranker is a permutation, and the model
must \emph{emit} that permutation as tokens: concretely, the $N$ ordinal values that
name the items in ranked order. Ranking is therefore an unusually
\emph{format-specialized} generation problem: the output alphabet is
$\{1,\dots,N\}$, the output length is known before decoding begins, and every value
must appear exactly once. It is this structure that determines which decoding
strategies are available, and it is the axis along which we organize both the prior
work and our contribution.

Autoregressive decoding spends one sequential forward pass over the KV cache
per emitted token, so decoding the $N$ ordinals uses $O(N \cdot T)$ passes, where $T$
is the number of tokens per item identifier. This is a bottleneck of the \emph{decoder},
not of the model: the prefill that computes the candidate representations is already
parallel and already optimized in its computational complexity. As an example, in the experiments for this paper, a teacher
with an explicit reasoning trace reaches $1807$\,ms per request on an A100 GPU, of which
$28$\,ms is prefill. We specifically target \emph{late-stage re-ranking}, in which a
modest candidate set retrieved by an upstream system must be ordered by relevance under
strict latency constraints.

\paragraph{Decoding strategies.}
Because the ordinals are what must be decoded, the design space is a space of
\emph{decoders} over a fixed backbone. Autoregressive decoding emits them left to right,
one forward pass per token: exact and requiring no architectural change, but with a pass
count that grows with the slate. Parallel decoders, non-autoregressive rerankers and
multi-token predictors \citep{DBLP:conf/kdd/RenYWXWZ24}, and diffusion-based rerankers
\citep{DBLP:journals/corr/abs-2602-12528}, emit all positions at once and
refine them over several steps, reducing the pass count but not to one, and\rev{, unless followed by an assignment step,} without
guaranteeing that the emitted ordinals form a permutation. We ask whether the ordinals
can be decoded in a single pass:

\textbf{Can a ranker decode all $N$ ordinals in $O(1)$ forward passes and still recover
the ranking that an autoregressive LLM computes over hundreds of sequential passes?}

We answer yes, with \spd, which exploits a connection between ranking and optimal
assignment. The key insight is that an LLM's prefill hidden states, computed in a single
parallel forward pass over all candidates, already encode the comparative information
that the ordinals express, via the backbone's self-attention layers. Autoregressive
decoding re-serializes that information one token at a time; \spd instead decodes it in
place. A lightweight self-attention head turns the prefill states into an $N \times K$
item--position score matrix, and the Hungarian algorithm \citep{kuhn1955hungarian}
decodes all $N$ ordinals from it as the optimal bipartite assignment, with a valid
permutation guaranteed by construction.

\paragraph{Contributions.}
\begin{enumerate}
\item \textbf{A formulation of ordinal decoding as optimal assignment on hidden states.} We cast the emission of the $N$ item ordinals as maximum-weight bipartite matching over an $N \times K$ item--position score matrix whose weights are read directly off the model's hidden states rather than off the output logits. This is the formulation that makes exact combinatorial decoding possible: the Hungarian algorithm recovers the optimal permutation at inference, from one pass. Output validity is guaranteed by construction rather than learned.
\item \textbf{\spd, an $O(1)$-pass decoding strategy for generative ranking.} A single prefill over all candidates yields per-item hidden states; a lightweight self-attention head maps them to the score matrix; the Hungarian solver decodes the $N$ ordinals from it. Decoding therefore takes a constant number of forward passes, independent of $N$, removing the sequential decode dependency that dominates autoregressive ranking latency.
\item \textbf{An empirical study on proprietary data and \rev{four public benchmarks}.} On our proprietary dataset, \rev{where teacher and student share the same $0.6$B architecture, \spd is $3.1\times$ faster than the teacher without reasoning ($88$\,ms to $28$\,ms) at on-par quality. Against the teacher with reasoning the gain is \speedup, of which $20.5\times$ comes from removing reasoning tokens ($1807$\,ms to $88$\,ms).} \rev{On Amazon Beauty, MS MARCO, BEIR Quora and BEIR MS MARCO we compare against ONE-PASS, Language Ranker, PE-Rank, FIRST and DiffuRank, and against two controls trained on the same teacher: an autoregressive student and a score-and-sort head. These controls separate what single-pass decoding contributes (speed and guaranteed validity) from what distillation contributes (quality).} We further ablate \rev{the scoring head, the Sinkhorn objective, the training signal and backbone adaptation, and measure scaling with slate size.} \end{enumerate}

\section{Background and Related Work}
\label{sec:related}

\subsection{LLM-Based Ranking and Generative Retrieval}
\citet{DBLP:conf/emnlp/0001YMWRCYR23} introduced RankGPT, showing that LLMs can rerank generatively,
decoding permutations autoregressively with a sliding window.
\citet{DBLP:conf/nips/Tay00NBM000GSCM22} pioneered generative retrieval with the Differentiable Search Index,
encoding document identifiers directly in model parameters, and
\citet{DBLP:conf/iclr/CaoI0P21} showed autoregressive entity generation is viable for structured
retrieval. Most relevant to our work, \citet{DBLP:conf/emnlp/ReddyDXSSSJ24} proposed FIRST, which uses
first-decoded-token logits for ranking. This is the closest prior approach to
constant-pass decoding, but it reads the ordering out of the output token distribution
rather than out of the hidden states. \spd keeps the same output as all of these, the
item ordinals, and changes only how they are decoded.


\subsection{Decoding Strategies for Generative Ranking}
Beyond autoregressive decoding, several lines of work reduce the number of sequential
forward passes. We organize them as autoregressive (AR), single-token decoding (STD),
speculative/parallel decoding (SD), and single-forward-pass (SFP) methods.

\paragraph{Autoregressive listwise decoding.}
RankGPT~\citep{DBLP:conf/emnlp/0001YMWRCYR23} decodes permutations autoregressively with a sliding
window; RankVicuna and RankZephyr~\citep{DBLP:journals/corr/abs-2309-15088,DBLP:journals/corr/abs-2312-02724}
distill RankGPT-style permutations into open 7B models with progressive multi-pass
windows. These are the quality ceiling for listwise generation but require $O(L)$
passes for $L$ emitted tokens plus multi-round sliding windows~\citep{DBLP:conf/emnlp/0001YMWRCYR23}.

\paragraph{Single-token decoding (STD).}
FIRST~\citep{DBLP:conf/emnlp/ReddyDXSSSJ24} fine-tunes with a learning-to-rank loss and reranks from
the logits of only the first generated identifier, completing reranking in a single
pass. STD is the closest prior approach to constant-pass decoding, but it approximates
the ranking from first-token signal and degrades severely on tail
positions~\citep{DBLP:conf/www/DuSWZ26}. SPD keeps the same output, the item ordinals, but
reads the full permutation out of prefill hidden states rather than out of the
first-token distribution.

\paragraph{Speculative and parallel decoding.}
Speculative decoding~\citep{DBLP:conf/icml/LeviathanKM23} drafts with a small model and
verifies with the target; blockwise decoding~\citep{DBLP:conf/nips/SternSU18} predicts
multiple future tokens with extra heads; Medusa~\citep{DBLP:conf/icml/CaiLGPLCD24} adds parallel
decoding heads with tree-attention verification; EAGLE~\citep{DBLP:conf/icml/LiW0024},
PDM~\citep{DBLP:conf/acl/SantilliSPMMMR23}, and lookahead decoding~\citep{DBLP:conf/icml/FuBS024}
extend this family, while LayerSkip~\citep{DBLP:conf/acl/ElhoushiSLHWL0A24} reduces target-model
computation through early-exit inference. These methods reduce the cost of
left-to-right generation, but remain multi-round and their pass count scales with
output length. They are therefore orthogonal and potentially composable with \spd,
which replaces sequential generation with a single-pass assignment. Tree
attention~\citep{DBLP:journals/corr/abs-2308-04623,DBLP:conf/icml/CaiLGPLCD24} enables parallel verification of
multiple prefixes in one pass; ONE-PASS~\citep{DBLP:conf/www/DuSWZ26} uses it to pre-verify
subset combinations ($N! \to 2^N \to K\cdot 2^{N/K}$) and reconstruct full-ranking
probability with M\"obius inversion plus hierarchical self-attention, as a tuning-free
approximation to autoregression. SPD differs: it never enumerates prefixes and never
re-queries the LLM, the permutation is decoded once, from hidden states, via exact
assignment.

\paragraph{Efficient listwise variants.}
PE-Rank~\citep{DBLP:conf/www/LiuWWM25} compresses each passage to a single embedding token and
constrains decoding to identifier tokens, improving prefill and decode efficiency. This
is orthogonal to SPD: PE-Rank shortens the prefill input while SPD eliminates the
sequential decode. \rev{Language Ranker~\citep{DBLP:conf/nips/ZhangDZXWWL25} also reads LLM hidden
states with a lightweight head, but uses the head to pick the best of $M$ sampled
autoregressive rankings, so it still pays for $M$ full generations; \spd generates no tokens.}

\rev{Zero-shot prompting families (pointwise, pairwise, listwise, setwise), differentiable permutation learning, and distillation for ranking are discussed in Appendix~\ref{app:related}.}

\subsection{Hungarian Assignment in Deep Learning}
DETR \citep{DBLP:conf/eccv/CarionMSUKZ20} uses bipartite matching between predicted and ground-truth
detections for set-based loss computation. \citet{DBLP:journals/corr/abs-2002-08676} provided
theoretical foundations for learning with differentiable perturbed combinatorial
optimizers. In mechanism design, \citet{DBLP:conf/icml/Duetting0NPR19} learn optimal multi-item
auctions via neural networks, and for ad allocation \citet{DBLP:journals/corr/abs-2506-05685} introduced
non-autoregressive generative auctions with position-query cross-attention. Critically,
we use Hungarian matching \emph{not} for loss computation as in DETR, but as the
inference-time decoder that replaces autoregressive generation. Similarly, DiffuRank~\citep{DBLP:journals/corr/abs-2602-12528} applies the Hungarian algorithm to \rev{the token distributions of masked rank slots in an 8B diffusion LM}. We instead apply it to \rev{a score matrix read off the hidden states of an autoregressive LLM by a learned head}, as the decoder itself, so the permutation constraint is enforced during decoding rather than restored afterwards.

Despite this breadth of prior work, no existing method decodes a complete ranking from an
LLM backbone in a constant number of forward passes, at real-time latency, with
near-lossless quality and a permutation guaranteed by construction.

\section{Problem Formulation}
\label{sec:formulation}

\subsection{Notation and Setup}
Let $u$ denote a user context and $C = \{c_1, \dots, c_N\}$ a slate of $N$ candidate
items retrieved by an upstream system. The re-ranking task is to produce a permutation
$\pi \in \Perm_N$ that orders items by relevance. An autoregressive teacher generates
$\pi$ via $T$ sequential decode passes, incurring $O(N \cdot T)$ latency. We seek a
student $f_\theta$ that decodes the same $N$ ordinals, and hence the same $\pi$, in
$O(1)$ decode passes, i.e.\ a single forward pass.

\paragraph{Latency decomposition.}
For an autoregressive ranker, inference computation is dominated by the decode phase: each
token requires a full forward pass over the KV-cache. The following examples are from the experiments in this paper. With reasoning enabled
$T \approx 768$ tokens; without reasoning $T \approx 39$. In both cases the prefill
phase, which processes all $N$ candidates in parallel, takes a fixed
$t_{\mathrm{prefill}} \approx 28$\,ms. Our key observation: if the ranking can be
extracted from the prefill hidden states alone, the $O(T)$ \emph{sequential} decode computational complexity
collapses to $O(1)$, the ordinals are still decoded, but not one at a time.

\subsection{Decoding a Ranking as Optimal Assignment}
Define a bipartite graph $G = (I \cup J, E)$ where $I = \{1,\dots,N\}$ indexes items
and $J = \{1,\dots,K\}$ indexes rank positions. Let $\Mmat \in \Reals^{N \times K}$ be a
score matrix in which $\Mmat_{ij}$ is the affinity of item $i$ for rank position $j$.
The optimal ranking is the maximum-weight bipartite matching
\begin{equation}
  \pi^{*} \;=\; \argmax_{\pi \in \Perm_N} \; \sum_{i=1}^{N} \Mmat_{i,\pi(i)}.
  \label{eq:lap}
\end{equation}
Equation~\eqref{eq:lap} is the linear assignment problem (LAP), solvable exactly in
$O(N^3)$ by the Hungarian algorithm \citep{kuhn1955hungarian}, or by the
Jonker--Volgenant variant (LAPJV) \citep{DBLP:journals/computing/JonkerV87} which achieves $O(N^2)$
average-case complexity via shortest augmenting paths.

\paragraph{Complexity comparison.}
For $N = 50$, the autoregressive teacher requires $T = 768$ sequential forward
passes. The Hungarian solver requires $O(N^3) \approx 1.25 \times 10^{5}$ arithmetic
operations, roughly $8\,\mu$s on CPU. The full \spd pipeline (one prefill plus one
Hungarian solve, i.e.\ $O(1)$ passes for all $N$ ordinals) is therefore bounded by the
computational complexity of prefill, making the combinatorial optimization effectively free
(Section~\ref{sec:solver}).

\subsection{Differentiable Relaxation via Sinkhorn}
The Hungarian algorithm is non-differentiable: it returns a discrete permutation. For
gradient-based training we require a continuous relaxation. We adopt the Sinkhorn
operator \citep{DBLP:conf/nips/Cuturi13,DBLP:conf/iclr/MenaBLS18}. When $N \neq K$, we
normalize the rows and columns to satisfy the corresponding prescribed marginal
constraints; in the square case $N=K$, this reduces to the usual doubly-stochastic
constraint over the Birkhoff polytope. In the $N=K$ case:
\begin{equation}
  \Smat^{(0)} = \exp(\Mmat / \tau), \qquad
  \Smat^{(l+1)} = \mathbf{D}_r^{-1} \, \Smat^{(l)} \, \mathbf{D}_c^{-1},
  \label{eq:sinkhorn}
\end{equation}
where $\mathbf{D}_r$ and $\mathbf{D}_c$ are diagonal normalization matrices enforcing
unit row and column sums, respectively, with $\mathbf{D}_c$ computed after the row
normalization. The temperature $\tau > 0$ controls relaxation sharpness
($\tau \to 0$ recovers a permutation matrix).

\paragraph{Training objective.}
Given a teacher permutation matrix $\Pteach \in \{0,1\}^{N \times K}$, the Sinkhorn
cross-entropy loss is
\begin{equation}
  \mathcal{L} \;=\; -\sum_{i,j} \Pteach_{ij} \, \log \Smat^{(L_S)}_{ij}.
  \label{eq:loss}
\end{equation}
This provides smooth gradients to $\Mmat_\theta$ while driving the student's output
toward a valid permutation. At inference we bypass Sinkhorn entirely and apply the
Hungarian algorithm to $\Mmat$ directly, guaranteeing an exact permutation.

\section{Method: \spd}
\label{sec:method}

\subsection{Architecture Overview}
\spd comprises three components applied in sequence (Figure~\ref{fig:arch}).
Figure~\ref{fig:arch} depicts the winning configuration (self-attention head, $L=2$,
LoRA-adapted backbone); alternative head variants are ablated in
Section~\ref{sec:ablation-head}.

\begin{figure}[t]
  \centering
  \input{figures/fig_architecture.tex}
  \caption{\spd architecture. A single prefill pass over all $N$ candidates yields
  per-item hidden states, which the self-attention head maps to an $N \times K$ score
  matrix. The Hungarian algorithm decodes all $N$ item ordinals from that matrix in one
  shot: $O(1)$ forward passes, no sequential decode. End-to-end latency is $28$\,ms.}
  \label{fig:arch}
\end{figure}

\paragraph{Backbone (prefill-only, LoRA-adapted).}
The backbone is a $0.6$B-parameter decoder-only transformer ($28$ layers, hidden
dimension $1024$, $16$ attention heads) pretrained for language modeling and fine-tuned
for ranking. It processes the full prompt containing the user context and all $N$
candidates in a single forward pass. We gather the hidden state at each candidate's
readout position (the final token of its description), yielding
$\{\hvec_1, \dots, \hvec_N\} \in \Reals^{N \times D}$.

\paragraph{Scoring head (self-attention, $L=2$).}
A lightweight transformer encoder with $L = 2$ self-attention layers operates over the
$N$ candidate vectors, allowing each item to attend to all others and thereby perform
explicit pairwise comparison. A linear projection maps each refined representation to
$K$ rank-position scores, producing $\Mmat \in \Reals^{N \times K}$.

\paragraph{Assignment decoder (Hungarian).}
Unlike DETR \citep{DBLP:conf/eccv/CarionMSUKZ20}, which uses Hungarian matching during training for
loss computation, we apply the Hungarian algorithm at inference as the decoding
mechanism itself. Applied to $\Mmat$ it yields the optimal one-to-one assignment: a
valid permutation decoded in one shot. The Hungarian solver is thus \spd's decoder in
the same sense that a sampling loop is an autoregressive decoder, it is the component
that turns model outputs into the $N$ emitted ordinals. During training,
Equation~\eqref{eq:loss} supplies gradients.

\rev{\paragraph{Fewer positions than items ($K<N$).} The score matrix need not be square.
With $K<N$ the rectangular Hungarian assigns $K$ distinct items to the $K$ positions, and
the remaining $N-K$ items follow, ordered by their best position score $\max_j \Mmat_{ij}$.
The loss is computed on the teacher's top-$K$ positions only. This \emph{top-$K$ head}
costs the same single pass, and it is the better choice when the teacher only ranks the
head of the list reliably (Section~\ref{sec:scaling}). With $K>N$, the solver fills $N$
of the $K$ positions and the ranking is read off in position order.}

\subsection{Scoring Head Variants}
\label{sec:heads}
We study three head architectures to isolate the contribution of cross-item comparison
(Figure~\ref{fig:heads} and Table~\ref{tab:heads} \rev{in Appendix~\ref{app:related}}).

\subsection{Training: Offline Teacher Ranking Distillation}
Training proceeds in two phases.

\textbf{Phase 1 (offline, one-time).} The teacher, a well-trained autoregressive
generative ranking model, decodes a ranking for every training slate via greedy
decoding in direct-output mode (no chain-of-thought). These target permutations are
computed once and stored. \rev{On the internal dataset, we} discard the small number of teacher outputs containing
repeated or missing ordinals. \rev{On the public benchmarks, the off-the-shelf 32B teacher
rarely emits a complete permutation: it names $21$--$27$ of the $50$ items and completes
only $0.8$--$1.3\%$ of slates. We complete its output by appending the missing items in
input order, the same parser we use to score every autoregressive ranker
(Appendix~\ref{app:completeness}). The tail of these targets is therefore input order, not
a teacher judgment.}

\textbf{Phase 2 (student training).} The student minimizes
Equation~\eqref{eq:loss} against the pre-computed teacher permutations. There is no
online interaction between teacher and student; the labels are fixed. This follows from
the decoding strategy: \spd's decoder has no intermediate state and no rollout, $\Mmat$
is conditioned only on the prompt, and the Hungarian solver is a deterministic function of
$\Mmat$, so the train-time and inference-time input distributions are identical by
construction. There is no exposure bias for on-policy distillation to correct and no
student trajectory for the teacher to score, and Equation~\eqref{eq:loss} is already a
whole-permutation objective rather than a per-token surrogate. One offline teacher pass
over the training set therefore suffices.

\subsection{Backbone Adaptation via LoRA}
\label{sec:lora}
We apply Low-Rank Adaptation \citep[LoRA,][]{DBLP:conf/iclr/HuSWALWWC22} with rank $64$ to all linear
projections in the backbone. This allows the backbone to reshape its hidden states to
encode ranking-relevant information while preserving pretrained knowledge. As
Section~\ref{sec:ablation-signal} shows, this adaptation is the enabler:
without it, even rich teacher signals cannot be absorbed by the scoring head.

\section{Experiments}
\label{sec:experiments}

\subsection{Setup}
\begin{description}
    \item[Internal dataset.] A proprietary re-ranking dataset ($\approx 189$K training slates, $236$K test requests), with up to $150$
      items per slate. \textbf{Teacher:} a well-trained autoregressive generative ranking model
($0.6$B-parameter decoder-only transformer, fine-tuned for ranking), evaluated both with
and without an explicit reasoning trace.
    \item[Amazon Beauty dataset.] The ``beauty'' products subset of the 2014 Amazon product
      dataset~\citep{DBLP:conf/www/HeM16}. Slates comprise 50 candidate items to rank, with 21,245 training and 1,118 evaluation slates. \textbf{Teacher:} 32B Qwen model used for ranking. We also test a 0.6B Qwen model for comparison.
    \item[\rev{Public reranking benchmarks.}] \rev{Three passage and question reranking sets in the same format as Amazon Beauty:
      $50$ candidates and one relevant item per slate, $1{,}118$ evaluation slates each.
      \emph{MS MARCO C-50} uses MS MARCO~\citep{DBLP:conf/nips/NguyenRSGTMD16} dev queries with Contriever~\citep{DBLP:journals/tmlr/IzacardCHRBJG22} top-$50$
      candidates; \emph{BEIR Quora} and \emph{BEIR MS MARCO}~\citep{DBLP:conf/nips/Thakur0RSG21} use the BM25~\citep{DBLP:journals/ftir/RobertsonZ09} top-$49$ hard
      negatives from the full corpus. Train and evaluation queries are disjoint, and we
      check for leakage (Appendix~\ref{app:data}). \textbf{Teacher:} Qwen3-32B~\citep{DBLP:journals/corr/abs-2505-09388}
      without reasoning, greedy, on all four public sets.}
  \end{description}

\textbf{Student.} The same $0.6$B architecture, adapted via LoRA ($r = 64$). Trained via \textsc{sft} from the respective teacher for each dataset.

\rev{\textbf{Baselines.} We run five efficient rerankers on the same slates, prompts, metric
code and teacher: ONE-PASS and Language Ranker on frozen Qwen3-0.6B and 32B backbones;
PE-Rank on a fully fine-tuned Qwen3-0.6B; FIRST (7B) with its released targets and retrained
on our teacher; and DiffuRank (LLaDA-8B), zero-shot and fine-tuned on our teacher
(Appendix~\ref{app:baselines}). \textbf{Controls.} Two models share \spd's backbone, LoRA, data,
step budget and teacher, and change only the decoder: an \emph{AR student} decoded greedily,
and a \emph{score+sort} head with one scalar per item, trained with a Plackett--Luce loss and
ranked by sorting.}

\textbf{Metrics.} AUC, Recall@$\{1,10\}$, NDCG@$1$\rev{; with one relevant item NDCG@1 equals Recall@1, so we omit it on the public sets.}
\textbf{Hardware.} NVIDIA A100 80GB; LAPJV solver on CPU. \rev{Latency is p50 per slate at batch size 1.}

\subsection{Main Results}
\label{sec:main-results}

  \begin{table}[t]
    \centering \small
    \begin{tabular}{@{}lrrrcccc@{}}
      \toprule
      \textbf{Model (0.6B) } & \textbf{Latency} & \textbf{Speed vs}  & \textbf{Decode} &
      \textbf{AUC} & \textbf{Recall@1} & \textbf{Recall@10} & \textbf{NDCG@1}\\
                     &                  & \textbf{no think AR} & \textbf{passes} & & & & \\
      \midrule
      \multicolumn{8}{@{}l}{\textbf{Internal Dataset}}\\
      \addlinespace[2pt]
      Teacher (with reasoning) & $1807$\,ms & \rev{$0.05\times$} & $768$ & $0.5911$ & $0.1634$ & $0.7877$ & $0.1776$\\
      Teacher (no reasoning)   & $88$\,ms   & \rev{$1.0\times$} & $39$  & $\mathbf{0.5912}$ & $0.1635$ & $0.7892$ &
  $0.1779$\\
      \addlinespace[2pt]
      \spd (ours)             & $\mathbf{28}$\,\textbf{ms} & \rev{$\mathbf{3.1}\boldsymbol{\times}$} & $\mathbf{1}$ &
        $0.5907$ & $\mathbf{0.1652}$ & $\mathbf{0.7912}$ & $\mathbf{0.1791}$\\
      \bottomrule
    \end{tabular}
    \caption{\rev{Results on the internal dataset, where teacher and student are both $0.6$B.} Every row decodes the same output: the $N$ item ordinals
    naming the ranked slate. \textbf{Decode passes} counts the sequential forward passes
    required to do so; for autoregressive teachers under greedy decoding this equals the
    number of emitted tokens, while \spd decodes the ordinals out of the prefill pass
    itself, so its count is constant in $N$. Speed-up is relative to the teacher \rev{without reasoning}.}
    \label{tab:main}
  \end{table}

We present the main results and a series of ablations below; note that ablations were performed on successive datasets. Table~\ref{tab:main} compares \spd against both teacher configurations. Removing the
reasoning trace reduces teacher latency from $1807$\,ms to $88$\,ms, a $20\times$
reduction in decode passes, at essentially no reduction in quality, confirming that reasoning
tokens are not load-bearing for ranking. \spd decodes the same ordinals in a single
pass, reaching $28$\,ms: $64\times$ faster than the reasoning teacher and
$3.1\times$ faster than the no-reasoning teacher. Note what the second comparison
isolates. The no-reasoning teacher has already stripped every non-essential token and
decodes only the ranking itself\rev{, and it has the same $0.6$B size as \spd}; the remaining $3.1\times$ is attributable purely to the
decoding strategy, not to shorter output \rev{or a smaller model}. \rev{We therefore treat $3.1\times$, not $64\times$, as the decoding speed-up on this dataset.} Quality is lossless on
list-level metrics: \spd is indistinguishable from both teachers on NDCG@1 ($0.1791$ \spd vs.\ $0.1776$ /
$0.1779$) and Recall@1 ($0.1652$ \spd vs.\ $0.1634$ / $0.1635$). AUC is the metric where \spd retains $99.9\%$ of the teacher's quality. We attribute the list-level gains to two factors: the Hungarian assignment enforces a globally consistent one-to-one ordering that autoregressive decoding only approximates, and distillation across the full training set averages out the teacher's per-slate errors
(Section~\ref{sec:capacity}). \rev{The controlled comparison in Section~\ref{sec:decode-vs-distill} suggests the second factor dominates.}

\begin{table}[t]
  \centering \scriptsize
  \setlength{\tabcolsep}{2.6pt}
  \begin{tabular}{@{}llrrrrrrrrrrrr@{}}
    \toprule
    & & \multicolumn{3}{c}{\textbf{Amazon Beauty}} & \multicolumn{3}{c}{\textbf{MS MARCO C-50}} & \multicolumn{3}{c}{\textbf{BEIR Quora}} & \multicolumn{3}{c}{\textbf{BEIR MS MARCO}}\\
    \cmidrule(lr){3-5}\cmidrule(lr){6-8}\cmidrule(lr){9-11}\cmidrule(l){12-14}
    \textbf{Method} & \textbf{Size} & AUC & R@1 & ms & AUC & R@1 & ms & AUC & R@1 & ms & AUC & R@1 & ms\\
    \midrule
    AR, zero-shot          & 0.6B & .512 & .023 & 921  & .510 & .024 & 4227 & .543 & .136 & 4479 & .505 & .016 & 633\\
    ONE-PASS               & 0.6B & .516 & .023 & 197  & .518 & .024 & 148  & .620 & .133 & 53   & .514 & .016 & 169\\
    Language Ranker        & 0.6B & .512 & .024 & 6.0s & .509 & .025 & 9.6s & .544 & .122 & 9.9s & .504 & .019 & 10.7s\\
    PE-Rank                & 0.6B & .753 & .198 & 173  & .683 & .078 & 174  & .968 & .646 & 174  & .854 & .220 & 173\\
    PE-Rank, 32B teacher & 0.6B & .655 & .117 & 1247 & .540 & .026 & 1263 & .933 & .609 & 1269 & .761 & .210 & 1259 \\
    AR student             & 0.6B & .616 & .135 & 1157 & .731 & .152 & 1215 & .957 & .714 & 1155 & .779 & .220 & 1204\\
    Score+sort             & 0.6B & .616 & .109 & 90    & .632 & .097 & 74    & .935 & .672 & 30   & .724 & .204 & 85\\
    \textbf{\spd} (ours)   & 0.6B & \textbf{.635} & \textbf{.135} & \textbf{84} & \textbf{.709} & \textbf{.074} & \textbf{76} & \textbf{.945} & \textbf{.731} & \textbf{28} & \textbf{.769} & \textbf{.191} & \textbf{85}\\
    \textbf{\spd}, top-10 head (ours) & 0.6B & \textbf{.685} & \textbf{.140} &  \textbf{89}  & \textbf{.793} & \textbf{.136} & \textbf{77}  & \textbf{.969} & \textbf{.726} & \textbf{29} & \textbf{.852} & \textbf{.214} & \textbf{77}\\
    \midrule
    FIRST, released targets & 7B  & .804 & .324 & 765  & .844 & .294 & 675  & .985 & .808 & 234  & .855 & .298 & 739\\
    FIRST, 32B teacher      & 7B  & .675 & .164 & 765  & .773 & .185 & 675   & .980 & .756 & 234   & .840 & .254 & 739\\
    DiffuRank, pointwise    & 8B  & .663 & .086 & 1933 & .791 & .177 & 1252 & .972 & .694 & 639  & .819 & .182 & 1388\\
    DiffuRank, Perm-Assign  & 8B  & .665 & .150 & 712  & .783 & .163 & 664  & .974 & .711 & 186  & .694 & .223 & 732\\
    ONE-PASS                & 32B & .629 & .137 & 2680 & .754 & .180 & 2389 & .967 & .705 & 738  & .802 & .217 & 2703\\
    Language Ranker         & 32B & .570 & .079 & 66s  & .732 & .186 & 46s  & .961 & .707 & 23s  & .776 & .220 & 41s\\
    AR teacher, no reasoning & 32B & .629 & .138 & 5075 & .737 & .182 & 3776 & .961 & .701 & 2455 & .786 & .216 & 3778\\
    \bottomrule
  \end{tabular}
  \caption{Public benchmarks: AUC, Recall@1 and p50 latency (ms unless marked s) on 1,118 evaluation slates of 50 candidates. Every trained model learns from the 32B teacher's order, except PE-Rank with relevance labels and FIRST with its released click or GPT-4 targets. Score+sort shares SPD's forward pass and backbone; its latency matches SPD's to within 6\,ms. The AR student on Beauty is shown at SPD's 500 steps (1,328 steps: .635 / .144). Training SPD on an 8k-slate subset of Beauty gives .617 / .132 at 113\,ms.}
  \label{tab:public}
\end{table}

\begin{figure}[t]
  \centering
  \includegraphics[width=\linewidth]{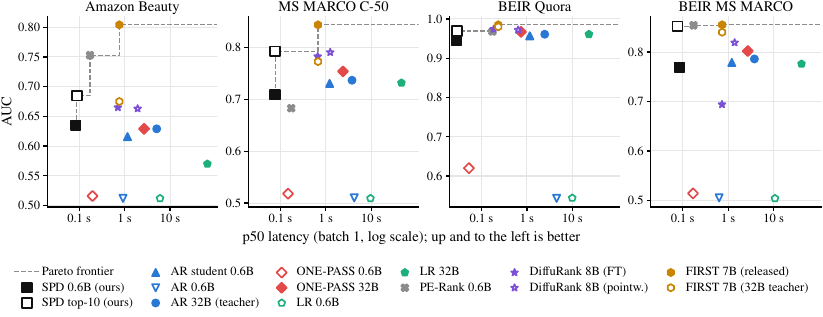}
  \vspace{-1.5em}
  \caption{\rev{Quality against latency on the public benchmarks (numbers from Table~\ref{tab:public}). The dashed line is the Pareto frontier: \spd is on it on every dataset. On the Pareto frontier, SPD represents the best tradeoff when accounting for model size and latency calculation settings (see below).}}
  \label{fig:quality-latency}
  \vspace{-1em}
\end{figure}

\rev{\subsection{Public Benchmarks and Efficient Rerankers}
\label{sec:public}
Table~\ref{tab:public} compares \spd with the efficient rerankers on the four public
benchmarks (plotted in Figure~\ref{fig:quality-latency}). At $0.6$B, ONE-PASS, Language Ranker and the
zero-shot autoregressive model are at chance (AUC $.50$--$.62$): a head on a frozen $0.6$B
backbone cannot recover the teacher's ranking. \spd is within $.03$ AUC of its 32B teacher on all
four datasets at $44$--$88\times$ lower latency, and has the best Recall@1 of any method on
BEIR Quora apart from FIRST. Its weak point is Recall@1 on MS MARCO C-50 ($.074$ against
${\approx}.18$ for the 32B rankers).

PE-Rank, fully fine-tuned on the relevance labels, is the strongest $0.6$B baseline: it
beats \spd on AUC on three datasets but trails it on MS MARCO C-50. The larger rerankers are better on quality. FIRST beats \spd everywhere,
even when retrained on our teacher, and fine-tuned DiffuRank beats it on three datasets, at
$12$--$13\times$ the parameters and $7$--$9\times$ the latency. The 32B ONE-PASS matches or exceeds the
teacher, and Language Ranker-32B comes close except on Beauty, at $0.7$--$66$\,s per slate. \spd's contribution is the quality it
reaches per millisecond and per parameter, not the highest quality.

The top-10 head ($K{=}10$ of $N{=}50$, Section~\ref{sec:method}) has higher AUC than the full head on
every dataset, and higher AUC than its own teacher. It never trains on the input-order tail of
the teacher's targets (Section~\ref{sec:scaling}). On TREC DL19/DL20~\citep{DBLP:journals/corr/abs-2003-07820,DBLP:conf/trec/CraswellMMYC20}, \spd reaches nDCG@10
$.544$/$.540$, against $.463$/$.495$ for BM25 and $.622$/$.598$ for its teacher (Appendix~\ref{app:trec}).}

\rev{\subsection{Separating Decoding from Distillation}
\label{sec:decode-vs-distill}
The AR student matches or slightly beats \spd on AUC on three datasets (MS MARCO C-50 $.731$
vs.\ $.709$, Quora $.957$ vs.\ $.945$, BEIR MS MARCO $.779$ vs.\ $.769$) and trails it on Beauty
($.616$ vs.\ $.635$). Both agree with the teacher's order to the same degree (Kendall $\tau$
$.38$--$.46$ vs.\ $.36$--$.45$; Appendix~\ref{app:fidelity}). The AR student is however
$14$--$41\times$ slower, and like its teacher it rarely finishes a ranking: only
$2.7$--$7.4\%$ of its outputs are complete permutations. The quality of \spd therefore comes
from distillation, and single-pass decoding supplies the speed and the guaranteed valid
output. Over the 32B teacher with reasoning, removing reasoning gives $6$--$10\times$,
distilling to $0.6$B while keeping autoregressive decoding gives $2$--$4\times$, and single-pass
decoding gives a further $14$--$41\times$ (Table~\ref{tab:decomp}, Appendix~\ref{app:baselines}).

Score+sort tests whether the assignment formulation matters, given a single pass. \spd has
higher AUC ($+.010$ to $+.077$) and Recall@10 (e.g.\ $.602$ vs.\ $.401$ on MS MARCO C-50) on
all four datasets, while score+sort has higher Recall@1 on the two MS MARCO sets: the
item--position objective buys whole-list quality rather than top-1 accuracy\rev{, as the sorted linear probe already suggested on the internal dataset (Section~\ref{sec:ablation-head})}.
At inference, the trained score matrix can also be decoded without Hungarian, by sorting
items on their expected position or by per-row argmax. This raises AUC by $.027$ on Quora and $.085$ on
BEIR MS MARCO with expected position, since AUC only rewards where the
relevant item lands, but Hungarian reproduces the teacher's permutation more faithfully on
MS MARCO C-50 and BEIR MS MARCO. Hungarian is the right decoder for a faithful permutation;
expected-position sorting is a free alternative when only the top items matter.}

\subsection{Ablation: Training Signal \texorpdfstring{$\times$}{x} Backbone Adaptation}
\label{sec:ablation-signal}

\begin{figure}[t]
  \begin{minipage}[b]{0.40\linewidth}
  \centering \footnotesize \setlength{\tabcolsep}{3pt}
  \begin{tabular}{@{}clcc@{}}
    \toprule
    \textbf{Track} & \textbf{Signal} & \textbf{Backbone} & \textbf{AUC} \\
    \midrule
    1 & Clicks  & Frozen & $0.5187$ \\
    2 & Teacher & Frozen & $0.5573$ \\
    3 & Teacher & LoRA   & $\mathbf{0.5842}$\\
    \bottomrule
  \end{tabular}
  \captionof{table}{Signal richness and backbone adaptation give complementary gains: teacher
  rankings beat click labels on a frozen backbone (Track 2 $>$ 1), and LoRA ($r{=}64$) adds more (Track 3 $>$ 2).}
  \label{tab:tracks}
  \end{minipage}\hfill
  \begin{minipage}[b]{0.57\linewidth}
  \centering
  \includegraphics[width=\linewidth]{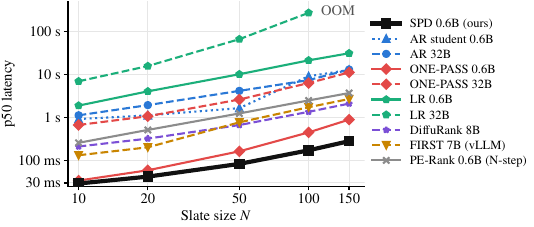}
  \captionof{figure}{\rev{p50 latency vs.\ slate size $N$. Solid: ${\le}0.6$B rankers; dashed: 7B/32B. 32B model is out of memory (OOM) for the largest slate size.}}
  \label{fig:latency-n}
  \end{minipage}
\end{figure}

With a frozen backbone, replacing binary click labels with full teacher rankings improves
AUC from $0.5187$ to $0.5573$ (Table~\ref{tab:tracks}); adapting the backbone raises it to $0.5842$.

\subsection{Ablation: Head Architecture}
\label{sec:ablation-head}
Table~\ref{tab:head-ablations} \rev{(Appendix~\ref{app:related})} shows our ablations on the three different head architectures described. The self-attention architecture described above performs best. Removing the self-attention layers and replacing them with a linear probe reduces the ranking performance, as does adding learned position-dependent embeddings. \rev{The linear probe ranks items by sorting its scores instead of solving the assignment, so it is also our score-and-sort ablation on this dataset.} On the backbone that is LoRA-adapted, the self-attention head is the strongest configuration, consistent with our
interpretation in Section~\ref{sec:capacity}.

\rev{\subsection{Ablations on the Public Benchmarks}
\label{sec:ablation-public}
Table~\ref{tab:ablation-public} repeats the training ablations on the public benchmarks and
adds the cell missing from Table~\ref{tab:tracks}, click labels with LoRA. Replacing Sinkhorn with a
per-item softmax cross-entropy lowers AUC on every dataset, most on MS MARCO C-50. Click
labels with LoRA are at chance on MS MARCO C-50 and $.12$--$.20$ AUC lower on Quora and
BEIR MS MARCO, and a frozen backbone is near chance except on Quora: as on the internal
dataset, distillation and backbone adaptation are both needed. Three further heads do not
beat the default head on both datasets we tried, and \spd carries over to
Llama-3.2-1B~\citep{DBLP:journals/corr/abs-2407-21783} and five other backbones (Appendix~\ref{app:heads}).

\begin{table}[t]
  \centering \small
  \setlength{\tabcolsep}{4pt}
  \begin{tabular}{@{}lcccc@{}}
    \toprule
    \textbf{Variant (AUC / R@1)} & \textbf{Beauty} & \textbf{MS MARCO C-50} & \textbf{BEIR Quora} & \textbf{BEIR MS MARCO}\\
    \midrule
    \spd (teacher, LoRA, Sinkhorn)  & \textbf{.635} / .135 & \textbf{.709} / \textbf{.074} & \textbf{.945} / \textbf{.731} & \textbf{.769} / \textbf{.191}\\
    No Sinkhorn                     & .619 / .129 & .574 / .029 & .936 / .694 & .753 / .169\\
    Click labels + LoRA             & .581 / \textbf{.155} & .510 / .023 & .820 / .662 & .570 / .158\\
    Teacher, frozen backbone        & .541 / .032 & .510 / .015 & .748 / .309 & .543 / .037\\
    \bottomrule
  \end{tabular}
  \caption{Training ablations on the public benchmarks, at \spd's step budgets.}
  \label{tab:ablation-public}
\end{table}}

\subsection{Solver Overhead}
\label{sec:solver}

\rev{On the internal dataset,} the backbone prefill accounts for
essentially all of the end-to-end latency (\rev{$27.9$ of $28$\,ms}); the scoring head adds under $0.1$\,ms and the LAPJV solver
$0.008$\,ms, under $0.03\%$ of the total. This is the central efficiency argument:
although the linear assignment problem is $O(N^3)$ in the worst case, at
slate sizes of $N \le 50$ it resolves in microseconds on CPU, fully overlapped by the GPU
forward pass. \spd specifically targets this late-stage re-ranking regime rather than
retrieval-scale candidate pools. The apparent overhead of introducing a combinatorial
solver is therefore illusory in our target setting. More importantly, the decomposition shows that \spd has moved the bottleneck
from a \emph{sequential} dependency (autoregressive decode, which cannot be
parallelized) to a \emph{parallel} one (a single prefill over all candidates). Decoding
all $N$ ordinals has an impact of $0.008$\,ms of that budget. Any further latency reduction must
come from the backbone itself, via pruning, quantization, or early exit, rather than from
the ranking machinery, which is already effectively free.

\rev{\subsection{Scaling with Slate Size and Robustness}
\label{sec:scaling}
Figure~\ref{fig:latency-n} shows that this holds as the slate grows: \spd's latency is
prefill-only and scales roughly linearly in $N$, from $29$\,ms at $N{=}10$ to $281$\,ms at
$N{=}150$, whereas autoregressive rankers scale with the number of emitted labels. The assignment step stays under $3$\,ms; at $N{=}150$
\spd is $46\times$ faster than the 32B teacher and $45\times$ faster than the AR student.

On quality, we retrain on MS MARCO Contriever slates of $N{=}20$ and $N{=}100$. At $N{=}20$,
\spd reaches AUC $.732$ against the teacher's $.745$. At $N{=}100$ the full head falls to
chance ($.511$), and the cause is the target: the teacher names a median of $16$ of the
$100$ items, so ${\approx}84\%$ of every target is input-order filler, and \spd learns to
reproduce it. The top-10 head avoids the filler and reaches $.814$ against the
teacher's $.712$.

Shuffling the candidates (three random orders and the reversed order) moves \spd's AUC by at
most $.02$ on every dataset, as for the 32B teacher; Language Ranker-32B is the most sensitive
($-.077$ on MS MARCO C-50). No ranker, \spd included, moves by more than $.05$ AUC between the shortest and
the longest third of slates by length (Appendix~\ref{app:robustness}).}

\rev{\subsection{Discussion}\label{sec:analysis}\label{sec:capacity}}

\paragraph{The capacity gap.}
Click labels supply a simple gradient (``push clicked items up''), whereas teacher rankings
supervise the complete ordering. The language-modeling-optimized representations limit how
fully a head can exploit this richer signal; LoRA narrows this capacity gap with roughly $40$M
trainable parameters on a $600$M backbone (Track 3\rev{; frozen-backbone rows of Table~\ref{tab:ablation-public}}).

\paragraph{Structural guarantees.}
By construction \spd yields (i) $100\%$ valid outputs, since the Hungarian algorithm
always returns a valid permutation; the constraint is discharged inside the decoder
rather than repaired after it; (ii) no degenerate solutions, since the one-to-one
constraint forbids assigning multiple items to the same position; and (iii)
deterministic inference, with no sampling temperature or beam-search artifacts. These
properties are guaranteed rather than learned, and they account for part of the
list-level advantage over the autoregressive teacher, which can and does emit invalid
or repeated orderings. \rev{On the public benchmarks the off-the-shelf 32B teacher completes a
permutation on only $0.8$--$1.3\%$ of slates and the zero-shot $0.6$B model on none (Appendix~\ref{app:completeness}).}

\section{Conclusion}
We presented \spd, a format-specialized decoding strategy that decodes a complete
ranking, all $N$ item ordinals, in $O(1)$ forward passes, at $28$\,ms end-to-end. We
formalize the use of the Hungarian algorithm to decode a ranking directly from hidden
states. We empirically validate this approach on a proprietary dataset and \rev{four public
benchmarks}, finding near-lossless ranking quality compared to autoregressive
LLMs\rev{, $3.1\times$ faster than a same-size teacher without reasoning and $14$--$41\times$
faster than an autoregressive student distilled from the same teacher.} We profile the performance of the
different parts of the architecture and our systematic ablations decompose the
contributions between the backbone and the self-attention layers.

\subsubsection*{Acknowledgments}
The authors would like to thank Yufei Li for providing the upstream data assets and model checkpoints used in this work, and for helpful discussions on evaluation methodology.



\bibliographystyle{iclr2027_conference}
\bibliography{references}

\rev{
\clearpage
\appendix
\section{Additional Related Work and Head Variants}
\label{app:related}
}
\subsection*{Zero-Shot Prompting Taxonomy: Pointwise, Pairwise, Listwise, Setwise}
Zero-shot LLM ranking is commonly organized into four prompting
families~\citep{DBLP:conf/sigir/ZhuangZKZ24,DBLP:conf/naacl/QinJHZWYSLLMWB24}. \emph{Pointwise} methods score each
document independently: MonoT5 recasts relevance scoring as the generation of
binary target tokens~\citep{DBLP:conf/emnlp/NogueiraJPL20}, an approach also
used in general LLM evaluation via yes/no likelihoods~\citep{DBLP:journals/tmlr/LiangBLTSYZNWKN23},
or query likelihood $P(\text{query}\mid\text{doc})$~\citep{DBLP:conf/emnlp/SachanLJAYPZ22}. They cost
$O(N)$ calls and batch well but need logit access and calibration, and lag on
effectiveness. \emph{Pairwise} methods (PRP~\citep{DBLP:conf/naacl/QinJHZWYSLLMWB24}; origins in
duoBERT~\citep{DBLP:journals/corr/abs-1910-14424} and duoT5~\citep{DBLP:journals/corr/abs-2101-05667}) prompt with
document pairs and aggregate via AllPairs $O(N^2)$, sorting $O(N \log N)$, or
sliding-$K$ $O(N)$, with heap/bubble sort plus top-$k$ early stopping. They are the most
effective zero-shot family but the most expensive. \emph{Listwise} methods prompt with
a window of documents and generate the ranked label
list~\citep{DBLP:conf/emnlp/0001YMWRCYR23,DBLP:journals/corr/abs-2305-02156}, using bottom-up sliding windows over multiple
passes; they need generation only and no calibration, but are bottlenecked by context
length, generation cost, and sensitivity to initial order~\citep{DBLP:conf/sigir/ZhuangZKZ24}.
\emph{Setwise prompting}~\citep{DBLP:conf/sigir/ZhuangZKZ24} generalizes pairwise sorting to
best-of-$c$ comparisons per LLM call, reducing heap sort to $O(k \log_c N)$ and bubble
sort to $O(k \cdot N/(c-1))$, plus a logit-based listwise-likelihood variant that avoids
generating the label list. It gives the best effectiveness--efficiency trade-off among
zero-shot methods, but still issues many LLM calls, requires no training, and provides
no permutation guarantee. SPD occupies a different point: a trained, single-pass
decoder that emits a valid permutation by construction (Hungarian solver) rather than
accelerating a generate-and-sort loop.

\paragraph{Delineation.}
Unlike ONE-PASS, which verifies enumerated subset-prefixes through the LLM in one wide
pass, SPD never enumerates prefixes and never re-queries the LLM: the ranking is read
off hidden states once. Unlike Setwise, which accelerates zero-shot sorting by comparing
$c > 2$ documents per LLM call, SPD replaces the generate-and-sort loop entirely with a
learned head plus exact assignment.

\subsection*{Optimal Transport and Differentiable Permutation Learning}
Our training procedure builds on differentiable relaxations of discrete permutations.
\citet{DBLP:conf/nips/Cuturi13} introduced entropic regularization for efficient optimal
transport, providing the algorithmic foundation for the Sinkhorn operator, and
\citet{DBLP:conf/iclr/MenaBLS18} extended this to learning latent permutations with
Gumbel-Sinkhorn networks. Alternative differentiable sorting approaches include
SoftSort \citep{DBLP:conf/icml/PrilloE20} and differentiable sorting networks
\citep{DBLP:conf/icml/PetersenBKD21}. Our contribution is connecting these permutation-learning
tools to LLM hidden states.

\subsection*{Knowledge Distillation for Ranking}
\citet{DBLP:journals/corr/HintonVD15} introduced soft-target distillation. For ranking,
\citet{DBLP:journals/corr/abs-2010-02666} demonstrated cross-architecture distillation from
cross-encoders to bi-encoders, and \citet{DBLP:journals/corr/abs-2312-02724} distilled generative reranking
ability into smaller models. We differ in two ways: we distill full
permutations via a Sinkhorn cross-entropy objective, and our ablation reveals a
previously undocumented interaction between signal richness and backbone capacity.

\begin{figure}[t]
  \centering
  \input{figures/fig_heads.tex}
  \caption{Scoring head variants. (a) A linear probe scores each item independently; all cross-item context arises implicitly from the backbone's self-attention during prefill. (b) The self-attention head performs explicit item-to-item comparison before scoring. (c) The slot-query head inverts the formulation: $K$ learnable position embeddings cross-attend over items. On a frozen backbone (a) is competitive; once the backbone is LoRA-adapted, (b) is strongest.}
  \label{fig:heads}
\end{figure}

\begin{table}[b]
  \centering \small
  \begin{tabular}{@{}llr@{}}
    \toprule
    \textbf{Head} & \textbf{Key property} & \textbf{Params} \\
    \midrule
    Linear probe & Each item scored independently; all cross-item context & $550$K \\
                 & arises implicitly from backbone self-attention & \\
    \addlinespace[2pt]
    Self-attention ($L{=}2$) & Explicit pairwise comparison: each item attends & $4.2$M \\
                 & to all others before scoring & \\
    \addlinespace[2pt]
    Slot-query & Dual formulation: $K$ learnable position embeddings & $4.0$M \\
                 & cross-attend over items (``what fills rank 1?'') & \\
    \bottomrule
  \end{tabular}
  \caption{Scoring head variants. All operate on identical frozen or LoRA-adapted
  backbone hidden states.}
  \label{tab:heads}
\end{table}

\rev{
}
\begin{table}[h]
  \centering \small
  \begin{tabular}{@{}llr@{}}
    \toprule
    \textbf{Head} & \textbf{AUC} & \textbf{Recall@1} \\
    \midrule
    Self-attention, $L{=}2$ (\spd) & $\mathbf{0.5907}$& $\mathbf{0.1652}$\\
    Linear probe &  $0.5841$&  $0.1650$\\
    Slot-query & $0.5816$& $0.1579$\\
    \bottomrule
  \end{tabular}
  \caption{Ablation on the different scoring head architectures in the ads ranking dataset.}
  \label{tab:head-ablations}
\end{table}
\rev{
\section{Public Benchmark Construction}
\label{app:data}
All three public sets use the Amazon Beauty slate format: $50$ shuffled candidates labelled
A to AX, one relevant item, and the same prompt template with ``Query:'' followed by the
candidates.
\textbf{BEIR Quora} has no training split, so evaluation uses $1{,}118$ held-out test queries and
training uses the dev queries plus the remaining test queries ($13{,}456$ slates). When a
query has several relevant questions, one is chosen at random and the others are never used
as negatives. The $49$ negatives are the top BM25 hits over the full corpus ($k_1{=}0.9$,
$b{=}0.4$), skipping the query's own question.
\textbf{BEIR MS MARCO} uses the training qrels for training ($21{,}243$ slates) and the dev queries
for evaluation, with $49$ BM25 negatives from the $8.8$M-passage corpus.
\textbf{MS MARCO C-50} takes the Contriever top-$50$ for MS MARCO dev queries and splits them
into disjoint train and evaluation queries; $47$ of the $1{,}118$ evaluation slates have two or
three relevant passages.
On both BEIR sets we find no shared query IDs and no exact text matches across splits, and no
evaluation positive appears among the training candidates; on Quora we also find no
near-duplicate queries (Jaccard ${\ge}0.8$).

\section{Baseline Implementation}
\label{app:baselines}

\textbf{ONE-PASS.} Reimplemented from the paper: frozen backbone, $10$ subsets of $5$
candidates plus $10$ dual subsets, $\lambda{=}2$, learning rate $10^{-3}$, batch $16$, stopping
epoch chosen on a $10\%$ validation split. We train its head toward the 32B teacher's order;
at $0.6$B, training toward the backbone's own decode gives the same quality.
Its head is sized for $N{=}50$, so it cannot run at $N{=}20$ or $100$ without retraining.
\textbf{Language Ranker.} Frozen backbone, $M{=}20$ sampled rankings per slate, head under
$0.5$M parameters. Head type, readout layer, training label and loss were selected on a
Beauty validation split. We report the head trained toward the teacher; the click-trained
head is within $.02$ AUC on every dataset and backbone.
\textbf{PE-Rank.} Qwen3-0.6B fully fine-tuned with a two-layer MLP projector over a frozen
Qwen3-Embedding-0.6B passage encoder, without the stage-1 alignment, trained for $3$ epochs on
the relevant item (the top-1 prefix of ListMLE); the ranking is read from the first rank slot. The 32B-teacher row uses the paper's constrained decoding.
A reproduction with a frozen backbone and the paper's $N$-step constrained decoding stays at
chance (AUC $.49$--$.56$) and takes $1.24$\,s at $N{=}50$; Figure~\ref{fig:latency-n} times this
version.
\textbf{FIRST.} Zephyr-7B-$\beta$, fully fine-tuned. The ``released targets'' row uses the
released GPT-4-target model on the two MS MARCO sets and the same recipe trained on clicks or
relevance labels on Beauty and Quora. The ``32B teacher'' row retrains the same recipe on our
teacher's orders.
\textbf{DiffuRank.} LLaDA-1.5-8B with the paper's RoPE scaling. We add a ``Ranking:'' prefix
before the masked slots, without which Recall@1 is at chance. Perm-Assign is fine-tuned on
our teacher with LoRA ($r{=}16$) and a masked diffusion loss on the identifier tokens, for about
a third of the paper's budget, and decoded with Hungarian.
\textbf{Controls.} The AR student and score+sort use LoRA $r{=}64$ and \spd's data, learning rate
and step budgets ($500$ / $500$ / $841$ / $1{,}328$ steps on Beauty / MS MARCO C-50 / Quora /
BEIR MS MARCO).

\begin{table}[h]
  \centering \small
  \setlength{\tabcolsep}{4pt}
  \begin{tabular}{@{}lrrrr@{}}
    \toprule
    \textbf{p50 latency, $N{=}50$} & \textbf{Beauty} & \textbf{MS MARCO C-50} & \textbf{BEIR Quora} & \textbf{BEIR MS MARCO}\\
    \midrule
    Qwen3-32B, with reasoning     & $50.0$\,s & $24.0$\,s & $20.5$\,s & $24.0$\,s\\
    Qwen3-32B, no reasoning       & $5{,}075$\,ms & $3{,}776$\,ms & $2{,}455$\,ms & $3{,}778$\,ms\\
    AR student, 0.6B              & $1{,}157$\,ms & $1{,}215$\,ms & $1{,}155$\,ms & $1{,}204$\,ms\\
    \spd, 0.6B                    & $84$\,ms & $76$\,ms & $28$\,ms & $85$\,ms\\
    \midrule
    Removing reasoning            & $9.9\times$ & $6.4\times$ & $8.4\times$ & $6.4\times$\\
    32B $\to$ 0.6B, autoregressive & $4.4\times$ & $3.1\times$ & $2.1\times$ & $3.1\times$\\
    Single pass (\spd vs.\ AR student) & $13.8\times$ & $16.0\times$ & $41.3\times$ & $14.2\times$\\
    \bottomrule
  \end{tabular}
  \caption{Speed-up decomposition on the public benchmarks. The last row is the decoding
  speed-up at fixed size, data and teacher.}
  \label{tab:decomp}
\end{table}

\section{Completeness of Autoregressive Outputs}
\label{app:completeness}
The parser keeps the identifiers a model emits, in order, and appends the missing ones in
input order, so every output is scored as a full permutation and no slate is discarded.
Table~\ref{tab:completeness} shows how much of each output this fallback supplies. The
zero-shot Qwen3-0.6B model stops after about $26$ generated tokens on Beauty; on over $99\%$ of
slates, its final ranking is exactly the input order. These are the ``$26$
decode passes'' we reported for this model in the submitted version: the passes it spends before
stopping, not a complete ranking. The AR student completes $2.7$--$7.4\%$ of its outputs and names $16$--$24$
items. \spd, score+sort, ONE-PASS, FIRST, PE-Rank and DiffuRank Perm-Assign produce complete
permutations by construction.

\begin{table}[h]
  \centering \small
  \begin{tabular}{@{}lcccccc@{}}
    \toprule
    & \multicolumn{3}{c}{\textbf{Qwen3-0.6B, zero-shot}} & \multicolumn{3}{c}{\textbf{Qwen3-32B teacher}}\\
    \cmidrule(lr){2-4}\cmidrule(l){5-7}
    \textbf{Dataset} & complete & labels & tokens & complete & labels & tokens\\
    \midrule
    Amazon Beauty & $0.0\%$ & $0.1$ & $26.1$ & $0.9\%$ & $27.1$ & 77.9\\
    MS MARCO C-50 & $0.0\%$ & $2.6$ & $112.6$ & $1.1\%$ & $24.5$ & $63.8$\\
    MS MARCO C-20 ($N{=}20$)  & 0.8\% & 4.5 & 8.2   & 36.8\% & 16.9 & 19.0 \\
MS MARCO C-100 ($N{=}100$) & 0.0\% & 6.2 & 151.6 & 1.5\%  & 24.4 & 181.4 \\
    BEIR Quora    & $0.0\%$ & $3.0$ & $142.5$ & $0.8\%$ & $21.5$ & $58.7$\\
    BEIR MS MARCO & $0.0\%$ & $3.2$ & $93.3$  & $1.3\%$ & $24.7$ & $53.0$\\
    \bottomrule
  \end{tabular}
  \caption{Share of complete permutations, distinct labels emitted, and mean generated
  tokens, at $N{=}50$ unless marked.}
  \label{tab:completeness}
\end{table}

\section{Fidelity to the Teacher}
\label{app:fidelity}
Because the teacher names only $21$--$27$ items, a Kendall $\tau$ over all $50$ would be
inflated by agreement on the input-order tail. Table~\ref{tab:fidelity} reports $\tau$ on the
items the teacher emits, with the input order as a floor. \spd and the AR student agree with
the teacher to the same degree; the 32B rankers, which share the teacher's backbone, agree
more. Decoding \spd's score matrix with Hungarian gives a higher $\tau$ than expected-position
sorting or per-row argmax on MS MARCO C-50 and BEIR MS MARCO, and argmax is slightly higher on
Quora.

\begin{table}[h]
  \centering \small
  \begin{tabular}{@{}lcccc@{}}
    \toprule
    \textbf{Ranker} & \textbf{Beauty} & \textbf{MS MARCO C-50} & \textbf{BEIR Quora} & \textbf{BEIR MS MARCO}\\
    \midrule
    Input order (floor)     & .224 & .273 & .051 & .210\\
    \spd 0.6B               & .403 & .361 & .429 & .452\\
    AR student 0.6B         & .379 & .411 & .455 & .457\\
    Score+sort 0.6B         & .400  & .370 & .425   & .442\\
    FIRST 7B, 32B teacher   & .358 & .420 & .454 & .441\\
    DiffuRank Perm-Assign 8B & .451 & .471 & .493 & .449\\
    ONE-PASS 32B            & .447 & .560 & .570 & .568\\
    Language Ranker 32B     & .291 & .640 & .703 & .672\\
    \bottomrule
  \end{tabular}
  \caption{Kendall $\tau$ to the 32B teacher on the items it emits. The $0.6$B zero-shot,
  ONE-PASS and Language Ranker models are at the floor.}
  \label{tab:fidelity}
\end{table}

\section{Robustness to Candidate Order and Length}
\label{app:robustness}
We re-score each ranker on three random permutations and the reversed order of every slate
(Table~\ref{tab:robustness}). \spd's AUC moves by at most $.02$. Its rankings of the same slate
under different orders agree only in the head (cross-order $\tau$ $.01$--$.12$, against
$.08$--$.17$ for the teacher), since the tail it learned is the teacher's input-order filler.
Splitting slates into terciles by passage length, no ranker's AUC differs by more than $.05$
between the shortest and the longest tercile.

\begin{table}[h]
  \centering \footnotesize
  \setlength{\tabcolsep}{2pt}
  \begin{tabular}{@{}lcccc@{}}
    \toprule
    \textbf{AUC, orig.\ $\to$ shuffled} & \textbf{Beauty} & \textbf{MS MARCO C-50} & \textbf{BEIR Quora} & \textbf{BEIR MS MARCO}\\
    \midrule
    \spd 0.6B               & .618 $\to$ .598 & .705 $\to$ .700 & .946 $\to$ .945 & .768 $\to$ .772\\
    AR student 0.6B         & .616 $\to$ .623 & .731 $\to$ .715 & .957 $\to$ .959 & .779 $\to$ .779\\
    AR teacher 32B          & .629 $\to$ .622 & .737 $\to$ .733 & .961 $\to$ .960 & .786 $\to$ .784\\
    FIRST 7B, 32B teacher   & .675 $\to$ .657 & .773 $\to$ .766 & .980$\to$ .979 & .840 $\to$ .840\\
    DiffuRank Perm-Assign 8B & .665 $\to$ .646 & .783 $\to$ .775 & .974 $\to$ .972 & .694 $\to$ .700\\
    Language Ranker 32B     & .570 $\to$ .549 & .732 $\to$ .655 & .961 $\to$ .919 & .776 $\to$ .728\\
    \bottomrule
  \end{tabular}
  \caption{AUC in the original order and averaged over the four reorderings. The \spd row
  re-evaluates the trained checkpoints; on Beauty, the 8k-slate model.}
  \label{tab:robustness}
\end{table}

\section{Additional Heads and Backbones}
\label{app:heads}
Table~\ref{tab:heads-public} adds three heads on Beauty and MS MARCO C-50. On Beauty all of
them peak within $100$--$300$ steps and then flatten or decline; the MLP and position-query heads beat the
default head there but fall well below it on MS MARCO C-50, so no head is best on both. Table~\ref{tab:backbones} retrains \spd with the same recipe, tuned for
Qwen3-0.6B, on six other backbones. It carries over to Llama-3.2-1B and to the hybrid-attention
Qwen3.5 models. Agreement with the teacher grows with size up to 4B on every dataset, and AUC grows
on all but Beauty; at 4B, \spd matches or exceeds its teacher's AUC on every dataset except Beauty.

\begin{table}[h]
  \centering \small
  \begin{tabular}{@{}lcc@{}}
    \toprule
    \textbf{Head (AUC / R@1)} & \textbf{Beauty} & \textbf{MS MARCO C-50}\\
    \midrule
    Self-attention, $L{=}2$ (\spd) & .635 / .135 & .709 / .074\\
    Pointwise MLP              & .651 / .136 (peak .661) & .624 / .030\\
    Position-query             & .658 / .134 (peak .668) & .570 / .026\\
    Self-attention, $L{=}4$    & .607 / .138 (peak .634) & .709 / .084\\
    \bottomrule
  \end{tabular}
  \caption{Additional scoring heads at the last step.}
  \label{tab:heads-public}
\end{table}

\begin{table}[t]
\centering
\setlength{\tabcolsep}{3pt}
\resizebox{\linewidth}{!}{%
\begin{tabular}{l cc cc cc cc r}
\toprule
 & \multicolumn{2}{c}{\textbf{Beauty}} & \multicolumn{2}{c}{\textbf{MS MARCO C-50}} & \multicolumn{2}{c}{\textbf{BEIR Quora}} & \multicolumn{2}{c}{\textbf{BEIR MS MARCO}} & \\
\cmidrule(lr){2-3} \cmidrule(lr){4-5} \cmidrule(lr){6-7} \cmidrule(lr){8-9}
\textbf{Backbone} & AUC / R@1 & $\tau$ & AUC / R@1 & $\tau$ & AUC / R@1 & $\tau$ & AUC / R@1 & $\tau$ & p50 \\
\midrule
Qwen3-0.6B   & .635 / .135 & .403 & .709 / .074 & .361 & .945 / .731 & .429 & .769 / .191 & .452 & 83\,ms \\
Qwen3-1.7B   & .633 / .155 & .411 & .709 / .110 & .418 & .951 / .715 & .440 & .775 / .199 & .473 & 142\,ms \\
Qwen3-4B     & .623 / .148 & .422 & .739 / .155 & .466 & .967 / .738 & .468 & .788 / .210 & .499 & 307\,ms \\
Qwen3-8B     & .629 / .145 & .416 & .750 / .157 & .470 & .967 / .728 & .477 & .782 / .211 & .507 & 471\,ms \\
Llama-3.2-1B & .629 / .141 & .408 & .703 / .090 & .362 & .957 / .710 & .440 & .765 / .193 & .453 & 81\,ms \\
Qwen3.5-0.8B & .631 / .137 & .408 & .740 / .135 & .407 & .955 / .718 & .441 & .775 / .205 & .462 & 81\,ms \\
Qwen3.5-4B   & .628 / .151 & .416 & .751 / .168 & .476 & .961 / .723 & .468 & .788 / .219 & .506 & 290\,ms \\
\bottomrule
\end{tabular}}
\caption{SPD on other backbones, same 32B teacher (teacher AUC: Beauty .629, MS MARCO C-50 .737, BEIR Quora .961,
BEIR MS MARCO .786); all Beauty rows train on the same 21k-slate set. $\tau$: Kendall $\tau$ to the
teacher on the items it emits. Qwen3.5 uses hybrid linear/full attention. p50: per-slate latency on
synthetic Beauty slates of $N{=}50$, batch 1, A100-80GB.}
\label{tab:backbones}
\end{table}     

\section{TREC Deep Learning}
\label{app:trec}
Table~\ref{tab:trec} reports nDCG@10 with the standard \texttt{trec\_eval} convention (ideal
DCG over all judged passages; queries without a slate score $0$), reranking the BM25 top-$50$.

\begin{table}[h]
  \centering \small
  \begin{tabular}{@{}lcc@{}}
    \toprule
    \textbf{Method} & \textbf{DL19} & \textbf{DL20}\\
    \midrule
    BM25                           & .463 & .495\\
    AR zero-shot, 0.6B             & .297 & .265\\
    ONE-PASS, 0.6B / 32B           & .318 / .594 & .275 / .575\\
    Language Ranker, 0.6B / 32B    & .288 / .548 & .255 / .534\\
    DiffuRank, zero-shot / fine-tuned (8B) & .466 / .642 & .445 / .637\\
    AR teacher, 32B                & .622 & .598\\
    \spd, 0.6B                     & .544 & .540\\
    \bottomrule
  \end{tabular}
  \caption{TREC DL19 and DL20 nDCG@10, BM25 top-$50$.}
  \label{tab:trec}
\end{table}
}

\end{document}

%% file: figures/fig_architecture.tex
\resizebox{\linewidth}{!}{%
\begin{tikzpicture}[
  font=\scriptsize,
  box/.style    ={draw=black!65, rounded corners=2pt, align=center,
                  inner sep=3pt, fill=blue!5},
  bigbox/.style ={box, fill=blue!10},
  hid/.style    ={draw=black!65, circle, inner sep=0pt, minimum size=5mm,
                  fill=blue!18, font=\tiny},
  ar/.style     ={-{Latex[length=1.6mm,width=1.3mm]}, semithick, black!70},
  lbl/.style    ={font=\tiny, align=center, text=black!60},
]

\node[box, minimum height=13mm, text width=17mm] (prompt) at (0,0)
  {\textbf{Prompt}\\[2pt] user context\\ $+$ $N$ candidates};

\node[bigbox, minimum height=18mm, text width=25mm] (bb) at (3.1,0)
  {\textbf{LLM Backbone}\\[2pt] single forward pass\\ (prefill only)};
\node[lbl, below=1.5mm of bb] {0.6B params, LoRA-adapted};
\draw[ar] (prompt.east) -- (bb.west);

\node[hid] (h1) at (6.0, 0.85) {$h_1$};
\node[hid] (h2) at (6.0, 0.25) {$h_2$};
\node[font=\tiny] (hdots) at (6.0,-0.25) {$\vdots$};
\node[hid] (hN) at (6.0,-0.85) {$h_N$};
\node[lbl] at (6.0,-1.55) {per-item\\ hidden states};
\foreach \n in {h1,h2,hN}{ \draw[ar] (bb.east) -- (\n.west); }

\node[box, minimum height=16mm, text width=21mm] (head) at (8.7,0)
  {\textbf{Self-Attention}\\ \textbf{Head} ($L{=}2$)};
\node[lbl, below=1.5mm of head] {items attend to\\ all other items};
\foreach \n in {h1,h2,hN}{ \draw[ar] (\n.east) -- (head.west); }

\begin{scope}[shift={(11.0,-0.6)}]
  \foreach \r/\c/\v in {
    0/0/12, 0/1/70, 0/2/8,  0/3/10,
    1/0/78, 1/1/9,  1/2/7,  1/3/6,
    2/0/8,  2/1/14, 2/2/66, 2/3/12,
    3/0/6,  3/1/9,  3/2/13, 3/3/72}{
    \fill[blue!\v] (\c*0.28, \r*0.28) rectangle ++(0.28,0.28);
  }
  \draw[black!55, thin] (0,0) grid[step=0.28] (1.12,1.12);
  \node[lbl] at (0.56,-0.35) {Score Matrix\\ $\mathbf{M} \in \mathbb{R}^{N\times K}$};
\end{scope}
\draw[ar] (head.east) -- (10.95,0);

\node[box, minimum height=13mm, text width=19mm] (hung) at (13.6,0)
  {\textbf{Hungarian}\\ \textbf{Algorithm}\\[2pt] \tiny (LAPJV)};
\draw[ar] (12.2,0) -- (hung.west);

\node[box, minimum height=13mm, text width=17mm, fill=green!7] (out) at (16.3,0)
  {\textbf{Ranked List}\\[2pt] $[B, D, A, E, C]$\\[1pt] \tiny(item ordinals)};
\draw[ar] (hung.east) -- (out.west);

\draw[decorate, decoration={brace, amplitude=4pt, raise=1mm}, black!60]
  ($(prompt.north west)+(0,0.35)$) -- ($(out.north east)+(0,0.35)$)
  node[midway, above=3mm, font=\small\bfseries, text=black] {28\,ms total};
\node[font=\small\itshape, text=black!70] at (8.2,-2.5) {All $N$ ordinals decoded in one pass};

\end{tikzpicture}}

%% file: figures/fig_heads.tex
\resizebox{\linewidth}{!}{%
\begin{tikzpicture}[
  font=\scriptsize,
  hid/.style   ={draw=black!65, circle, inner sep=0pt, minimum size=5mm,
                 fill=blue!18, font=\tiny},
  op/.style    ={draw=black!65, rounded corners=2pt, align=center,
                 inner sep=2.5pt, fill=blue!6, font=\tiny},
  blk/.style   ={draw=black!55, rounded corners=3pt, fill=blue!4},
  ar/.style    ={-{Latex[length=1.3mm,width=1.1mm]}, thin, black!70},
  dar/.style   ={-{Latex[length=1.3mm,width=1.1mm]}, thin, black!70, dashed},
  aa/.style    ={{Latex[length=1.1mm]}-{Latex[length=1.1mm]}, thin, black!45},
  ttl/.style   ={font=\small\bfseries, align=center},
  capn/.style  ={font=\tiny, align=center, text=black!60},
]

\begin{scope}[shift={(0,0)}]
  \node[ttl] at (2.15,4.5) {(a) Linear Probe};
  \foreach \i/\x/\lab in {1/0.55/{$h_1$}, 2/2.15/{$h_2$}, 3/3.75/{$h_N$}}{
    \node[hid] (a\i) at (\x,3.7) {\lab};
  }
  \node[font=\tiny] at (2.95,3.7) {$\cdots$};
  \foreach \i/\x in {1/0.55, 2/2.15, 3/3.75}{
    \node[op, text width=12mm] (aL1\i) at (\x,2.6) {Linear\\$(D,512)$};
    \node[op, text width=12mm] (aL2\i) at (\x,1.5) {Linear\\$(512,K)$};
    \draw[ar] (a\i) -- (aL1\i);
    \draw[ar] (aL1\i) -- (aL2\i);
    \draw[ar] (aL2\i) -- (\x,0.25);
  }
  \node[capn, text width=32mm, anchor=north] at (2.15,0.15)
    {each item scored \emph{independently}\\ $\sim$550K params};
\end{scope}

\begin{scope}[shift={(5.3,0)}]
  \node[ttl] at (1.7,4.5) {(b) Self-Attention ($L{=}2$)};
  \foreach \i/\x/\lab in {1/0.4/{$h_1$}, 2/1.3/{$h_2$}, 3/3.0/{$h_N$}}{
    \node[hid] (b\i) at (\x,3.7) {\lab};
  }
  \node[font=\tiny] at (2.15,3.7) {$\cdots$};
  \node[blk, minimum width=34mm, minimum height=8mm] (bl1) at (1.7,2.75) {};
  \node[font=\tiny] at (1.7,3.05) {Self-Attention Layer 1};
  \foreach \x in {0.5,1.2,1.9,2.6,2.9}{ \node[hid, minimum size=2.6mm] at (\x,2.55) {}; }
  \draw[aa] (0.5,2.55) -- (1.2,2.55); \draw[aa] (1.2,2.55) -- (1.9,2.55);
  \draw[aa] (1.9,2.55) -- (2.6,2.55); \draw[aa] (0.5,2.55) to[bend right=28] (1.9,2.55);
  \draw[aa] (1.2,2.55) to[bend right=28] (2.6,2.55);
  \draw[aa] (0.5,2.55) to[bend right=34] (2.9,2.55);
  \node[blk, minimum width=34mm, minimum height=8mm] (bl2) at (1.7,1.55) {};
  \node[font=\tiny] at (1.7,1.85) {Self-Attention Layer 2};
  \foreach \x in {0.5,1.2,1.9,2.6,2.9}{ \node[hid, minimum size=2.6mm] at (\x,1.35) {}; }
  \draw[aa] (0.5,1.35) -- (1.2,1.35); \draw[aa] (1.2,1.35) -- (1.9,1.35);
  \draw[aa] (1.9,1.35) -- (2.6,1.35); \draw[aa] (0.5,1.35) to[bend right=28] (1.9,1.35);
  \draw[aa] (1.2,1.35) to[bend right=28] (2.6,1.35);
  \foreach \i in {1,2,3}{ \draw[ar] (b\i) -- (b\i |- bl1.north); }
  \draw[ar] (bl1.south) -- (bl2.north);
  \node[op, text width=13mm] (bL) at (1.7,0.72) {Linear\\$(512,K)$};
  \draw[ar] (bl2.south) -- (bL.north);
  \draw[ar] (bL.south) -- ++(0,-0.22);
  \node[capn, text width=32mm, anchor=north] at (1.7,0.15)
    {items compare to \emph{each other}\\ $\sim$4.2M params};
\end{scope}

\begin{scope}[shift={(10.6,0)}]
  \node[ttl] at (1.9,4.5) {(c) Slot-Query};
  \foreach \i/\y/\lab in {1/3.7/{$h_1$}, 2/3.0/{$h_2$}, 3/1.6/{$h_N$}}{
    \node[hid] (c\i) at (3.2,\y) {\lab};
  }
  \node[font=\tiny] at (3.2,2.3) {$\vdots$};
  \foreach \i/\y/\lab in {1/3.7/{$t_1$}, 2/3.0/{$t_2$}, 3/1.6/{$t_K$}}{
    \node[hid, fill=orange!22] (t\i) at (0.4,\y) {\lab};
  }
  \node[font=\tiny] at (0.4,2.3) {$\vdots$};
  \node[capn, text width=22mm] at (0.4,1.05) {learnable position\\ embeddings};
  \foreach \i in {1,2,3}{ \foreach \j in {1,2,3}{ \draw[dar] (t\i) -- (c\j); } }
  \node[font=\tiny, align=center] at (1.9,4.05) {cross-attention};
  \node[op, text width=17mm] (cdot) at (1.9,0.72) {dot product\\ $\to \mathbf{M}$};
  \draw[ar] (3.2,1.15) -- (cdot.east);
  \draw[ar] (cdot.south) -- ++(0,-0.22);
  \node[capn, text width=32mm, anchor=north] at (1.9,0.15)
    {positions \emph{query} items\\ $\sim$4.0M params};
\end{scope}

\node[capn, text width=60mm] at (7.2,4.95)
  {\textbf{Input (shared):} per-item hidden states $h_1,\dots,h_N$ from one prefill pass};

\node[draw=black!65, rounded corners=2pt, fill=green!7, font=\tiny, align=center,
      inner sep=3pt] (out) at (7.2,-0.95) {\textbf{Output:} score matrix $\mathbf{M}\,[N \times K]$
      $\to$ Sinkhorn (train) / Hungarian (eval)};

\foreach \x in {2.15,12.5}{ \draw[thin, black!70] (\x,-0.45) -- (\x,-0.60); }
\draw[thin, black!70] (2.15,-0.60) -- (12.5,-0.60);
\draw[ar] (7.2,-0.60) -- (out.north);

\end{tikzpicture}}